\documentclass[10pt,letterpaper]{article}
\usepackage{cogsci}
\cogscifinalcopy
\usepackage[utf8]{inputenc} %
\usepackage[T1]{fontenc}    %
\usepackage{microtype}
\usepackage{times,latexsym}
\usepackage{graphicx} \graphicspath{{figures/}}
\usepackage{amsmath,nicefrac}
\usepackage{acronym}
\usepackage[breaklinks,colorlinks,citecolor=black,bookmarks=true]{hyperref}
\usepackage{balance}
\usepackage{xspace}
\usepackage{setspace}
\usepackage[skip=3pt,font=small]{subcaption}
\usepackage[skip=3pt,font=small]{caption}
\usepackage[dvipsnames,svgnames,x11names,table]{xcolor}
\usepackage[capitalise,noabbrev,nameinlink]{cleveref}
\usepackage{booktabs,tabularx,colortbl,multirow,multicol,array,makecell,tabularray}
\usepackage{enumitem}
\usepackage[misc]{ifsym}
\usepackage{dblfloatfix}

\makeatletter
\DeclareRobustCommand\onedot{\futurelet\@let@token\@onedot}
\def\@onedot{\ifx\@let@token.\else.\null\fi\xspace}
\def\eg{\emph{e.g}\onedot}

\def\vs{\emph{vs}\onedot}

\makeatother

\makeatletter
\renewcommand{\paragraph}{%
    \@startsection{paragraph}{4}%
    {\z@}{0ex \@plus 0ex \@minus 0ex}{-1em}%
    {\normalfont\normalsize\bfseries}%
}
\makeatother

\usepackage[backend=biber,style=apa,natbib=true,annotation=false,backref=true]{biblatex}
\crefname{algorithm}{Alg.}{Algs.}
\Crefname{algocf}{Algorithm}{Algorithms}
\crefname{section}{Sec.}{Secs.}
\Crefname{section}{Section}{Sections}
\crefname{table}{Tab.}{Tabs.}
\Crefname{table}{Table}{Tables}
\crefname{figure}{Fig.}{Figs.}
\Crefname{figure}{Figure}{Figures}
\crefname{equation}{Eq.}{Eqs.}
\Crefname{equation}{Equation}{Equations}
\crefname{appendix}{Appx.}{Appxs.}
\Crefname{appendix}{Appendix}{Appendices}

\AtBeginDocument{
    \newcommand{\benchmark}{\textit{Overhang Tower}\xspace}
    
    \acrodef{ai}[AI]{Artificial Intelligence}
    \acrodef{cnn}[CNN]{Convolutional Neural Networks}
    \acrodef{rl}[RL]{Reinforcement Learning}
    \acrodef{voe}[VoE]{Violation-of-Expectation}
    \acrodef{ipe}[IPE]{Intuitive Physics Engine}
    \acrodef{ppo}[PPO]{Proximal Policy Optimization}
    \acrodef{sem}[SEM]{Standard Error of the Mean}
    \acrodef{com}[CoM]{Center of Mass}
}

\definecolor{tableHeaderGray}{HTML}{F2F2F2}
\definecolor{dataGreen}{HTML}{E2F0D9}
\definecolor{dataBlue}{HTML}{DDEBF7}
\definecolor{nsblue}{RGB}{51,102,204}     %
\definecolor{nsred}{RGB}{204,51,51}       %
\definecolor{nsgreen}{RGB}{0,153,102}     %
\definecolor{nsorange}{RGB}{255,153,51}   %
\definecolor{nspurp}{RGB}{102,51,153}     %

\title{Overhang Tower: Resource-Rational Adaptation in Sequential Physical Planning}

\author{%
    Ruihong Shen$^{1,2,3,4,5}$, Shiqian Li$^{2,1,4,5}$, and Yixin Zhu$^{1,2,4,5}$
    \vspace{3pt}\\\normalfont
    \small $^1$ School of Psychological and Cognitive Sciences, Peking University\quad
    \small $^2$ Institute for Artificial Intelligence, Peking University\\
    \small $^3$ School of EECS, Peking University\quad
    \small $^4$ State Key Laboratory of General Artificial Intelligence, Peking University\\
    \small $^5$ Beijing Key Laboratory of Behavior and Mental Health, Peking University\quad
    \small Project Website: \url{https://overhangtower.github.io}
    \vspace{-12pt}
}

\begin{document}
\maketitle

\begin{abstract}
Humans effortlessly navigate the physical world by predicting how objects behave under gravity and contact forces, yet how such judgments support sequential physical planning under resource constraints remains poorly understood.
Research on intuitive physics debates whether prediction relies on the \ac{ipe} \citep{battaglia2013simulation,zhang2016comparative} or fast, cue-based heuristics; separately, decision-making research debates deliberative lookahead \vs myopic strategies \citep{kahneman2011thinking}. These debates have proceeded in isolation, leaving the cognitive architecture of sequential physical planning underspecified.
How physical prediction mechanisms and planning strategies jointly adapt under limited cognitive resources remains an open question.
Here we show that humans exhibit a \textit{dual transition} under resource pressure, simultaneously shifting both physical prediction mechanism and planning strategy to match cognitive budget.
Using \benchmark, a construction task requiring participants to maximize horizontal overhang while maintaining stability, we find that \acs{ipe}-based simulation dominates early stages while \acs{cnn}-based visual heuristics prevail as complexity grows; concurrently, time pressure truncates deliberative lookahead, shifting planning toward shallower horizons---a dual transition unpredicted by prior single-mechanism accounts.
These findings reveal a hierarchical, resource-rational architecture that flexibly trades computational cost against predictive fidelity.
Our results unify two long-standing debates---simulation \vs heuristics and myopic \vs deliberative planning---as a dynamic repertoire reconfigured by cognitive budget.

\end{abstract}

\section{Introduction}

From infancy, humans develop a robust intuitive understanding of the physical world \citep{mccloskey1983intuitive,spelke2007core,piloto2022intuitive}, enabling not only passive predictions---such as whether a stack of boxes will topple---but also active planning of complex arrangements, like building a cantilevered shelf. This capacity for \textit{sequential physical planning}, the coordination of multiple interdependent actions to achieve a physical goal, is central to everyday cognition, yet its underlying mechanisms remain poorly understood.

\begin{figure}[t!]
    \centering
    \small
    \begin{subfigure}[t]{\linewidth}
        \includegraphics[clip,trim=0.2cm 0cm 0cm 0cm,width=\linewidth]{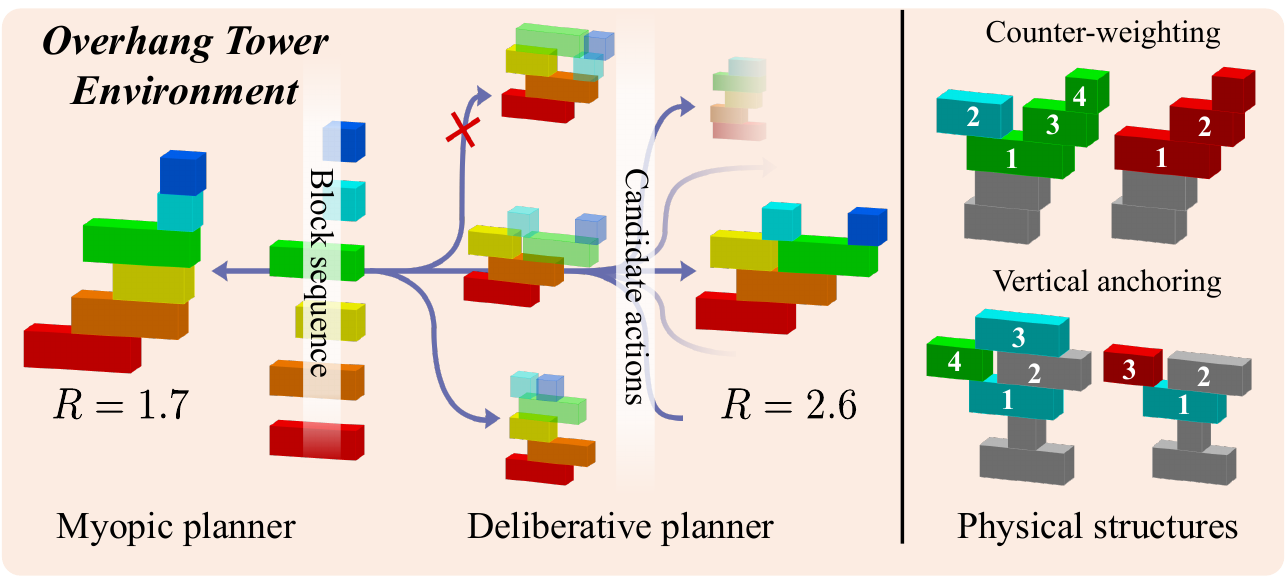}
        \caption{The \benchmark environment and task characteristics}
        \label{fig:intro_a}
    \end{subfigure}%
    \\%
    \begin{subfigure}[t]{0.304\linewidth}
        \includegraphics[width=\linewidth]{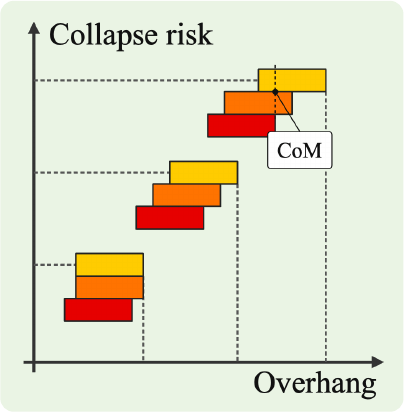}
        \captionsetup{justification=centering}
        \caption{Risk-reward\\trade-off}
        \label{fig:intro_b}
    \end{subfigure}%
    \hfill%
    \begin{subfigure}[t]{0.688\linewidth}
        \includegraphics[width=\linewidth]{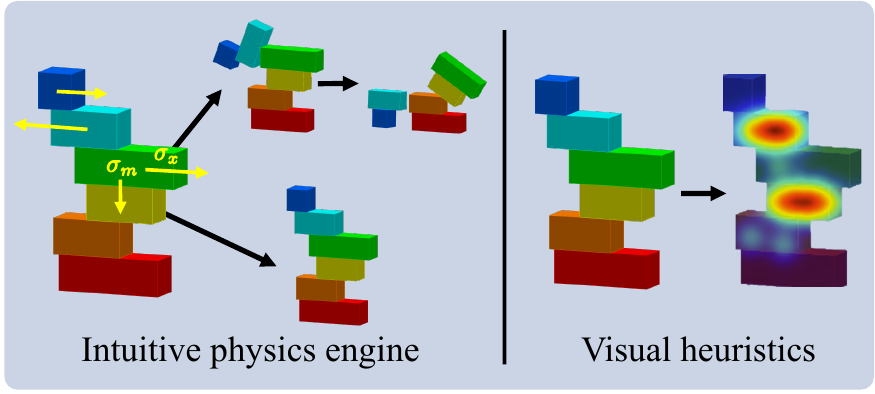}
        \caption{Physical predictions}
        \label{fig:intro_c}
    \end{subfigure}%
    \caption{\textbf{Overview of \benchmark and computational models.} (a) In \benchmark, participants construct a tower by placing blocks from a given sequence to maximize horizontal overhang while maintaining continuous stability. Due to \textit{resource-rational constraints} imposed by a large combinatorial search space, different planning strategies emerge. A myopic planner greedily seeks immediate overhang gains, often falling into local optima or causing collapse. In contrast, a deliberative planner looks ahead to build specific \textit{physical structures}---such as counter-weighting and vertical anchoring (numbers denote placement order)---that may sacrifice early reward but unlock higher long-term overhang ($R=2.6$ \vs $R=1.7$). (b) The task inherently involves a \textit{risk-reward trade-off}: extending blocks increases the overhang but shifts the \ac{com}, depleting the stability budget and making purely greedy approaches suboptimal. (c) To evaluate stability during planning, models rely on different \textit{physical prediction} mechanisms. The \ac{ipe} estimates stability via noisy probabilistic simulation of physical dynamics; alternatively, visual heuristics approximate stability through learned superficial visual patterns, bypassing explicit simulation.}
    \label{fig:intro}
\end{figure}

\begin{figure*}[t!]
    \centering
    \small
    \includegraphics[width=\linewidth]{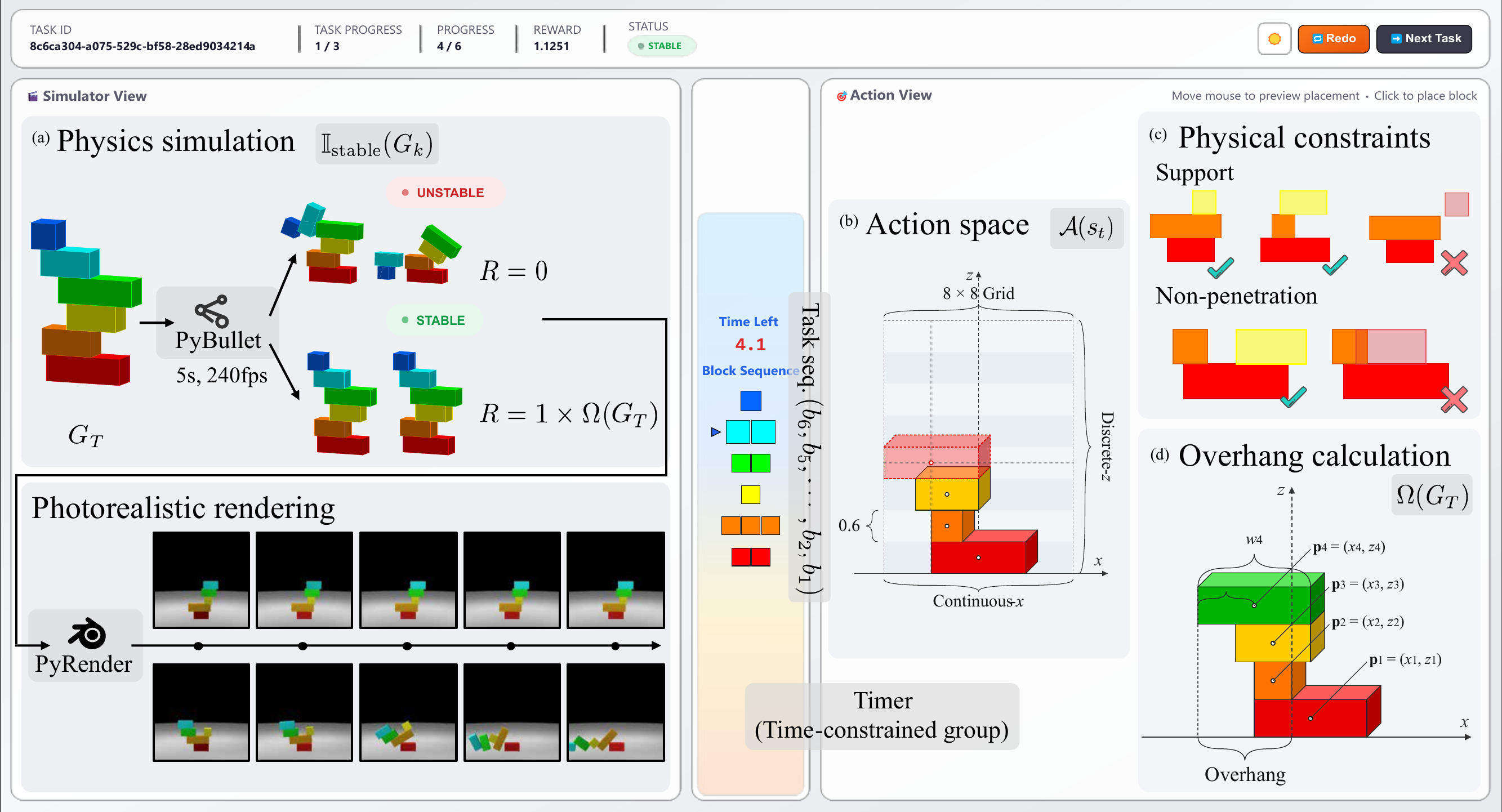}
    \caption{\textbf{Task interface and environment.} (a) Upon placement confirmation, the engine simulates dynamics to determine stability. Stable configurations yield a reward proportional to their overhang; any collapse results in zero reward. Simulated states are rendered into photorealistic visual feedback displaying the physical consequences of each placement. (b) Participants formulate actions within a hybrid spatial grid comprising continuous horizontal positioning and discrete height layers. (c) Placements are governed by non-penetration and support physical constraints; visual feedback instantly denotes action validity. (d) The overhang is derived from the structure's maximum lateral extension. The central panel displays the full block sequence for reference, alongside a timer indicating remaining time (time-constrained condition only).}
    \label{fig:stimuli}
\end{figure*}

Previous research on intuitive physics has primarily focused on \textit{passive, single-step judgments}: observers view a scene and predict a future outcome without intervening \citep{hamrick2011internal,battaglia2013simulation,zhang2016comparative,bear2021physion}. While these accounts successfully explain how humans judge stability from static images \citep{zhou2023mental,wang2024probabilistic,calabro2025cnn}, they leave a critical gap: how do physical prediction mechanisms guide \textit{sequential physical planning}, where agents must coordinate multiple interdependent decisions over time? Some recent interactive benchmarks introduce agency into physical reasoning, covering single-shot interventions \citep{bakhtin2019phyre,allen2020rapid} or multi-step interactions emphasizing action timing \citep{li2024iphyre,li2026neural}. However, these tasks typically frame physical planning as a \textit{binary satisfaction problem}---finding any successful causal chain to complete a goal---and none isolate the distinct computational challenges of \textit{sequential planning under constraints}: (i) early actions restrict future possibilities, demanding deliberative lookahead; (ii) cumulative risk must be managed to balance safety and reward; (iii) computational resources are insufficient to exhaustively evaluate all candidate sequences \citep{binder2025humans,mccarthy2020learning,huys2015interplay}.

To cope with such constraints, a natural hypothesis is that humans adaptively adjust the planning horizon, trading off speed against reward \citep{keramati2011speed,kuperwajs2025looking,snider2015prospective}: under time pressure, they rely on myopic, greedy action selection; when resources permit, they engage in deliberative lookahead, evaluating multi-step sequences before committing \citep{kahneman2011thinking}. While this dichotomy successfully explains behavior in abstract sequential decision tasks such as chess \citep{daw2011model,holding1989counting,russek2025time}, applying it to the physical domain introduces a unique complication: the nature of the underlying physical prediction mechanism itself remains contested. Computational accounts have modeled physical prediction either as probabilistic mental simulation---formalized in the \acf{ipe} framework \citep{hamrick2011internal,battaglia2013simulation,smith2013sources,ullman2017mind}---or as fast, cue-based heuristics that bypass explicit physics \citep{tversky1974judgment,sanborn2013reconciling,callaway2017discovering,calabro2025cnn}. Crucially, these two debates---myopic \vs deliberative planning strategies, and simulation \vs heuristics in physical prediction---have proceeded largely in isolation: planning research assumes a perfect oracle model of physics, while intuitive physics research focuses on passive judgments, neglecting how prediction mechanisms support sequential action.

This theoretical fragmentation leaves the cognitive architecture of sequential physical planning underspecified, raising two central questions: At the level of physical prediction, \textit{do humans simulate physical dynamics or rely on learned visual heuristics to approximate stability?} At the level of action selection, \textit{do humans employ fast, myopic strategies or engage in deliberative lookahead when constructing physical structures?}

We address both questions through the lens of \textit{resource-rationality} \citep{griffiths2015rational,callaway2018resource,callaway2022rational}. Recent work on passive judgment suggests simulation and heuristics constitute a dynamic repertoire, with costly simulation deployed only when cheaper heuristics prove insufficient \citep{li2025dual}. Extending this logic to sequential planning, we hypothesize a \textit{dual transition} under resource pressure: as construction progresses and scene complexity grows, participants should shift their physical prediction mechanism from costly \acs{ipe}-style simulation toward visual heuristics; simultaneously, as time pressure increases, their planning strategy should transition from deliberative lookahead toward myopic responding. Together, these constitute a hierarchical resource-rational architecture in which both physical prediction mechanism and planning strategy adapt dynamically to match cognitive budget to task demands.

To test this hypothesis, we introduce \benchmark, an interactive construction task in which participants place blocks to maximize horizontal overhang while maintaining continuous stability, as shown in \cref{fig:intro_a}. Unlike classic height-maximization stacking tasks \citep{groth2018shapestacks}, this objective enforces an explicit \textit{risk-reward trade-off} (\cref{fig:intro_b}): achieving high rewards requires placing blocks near stability limits, rendering the planner acutely sensitive to physical prediction errors. Greedy strategies may over-commit to stable but suboptimal placements, whereas deeper lookahead can justify early scaffolding that unlocks subsequent high-reward opportunities---making \benchmark particularly diagnostic for dissociating physical prediction mechanism from planning strategy.

To systematically vary cognitive resources, we manipulated \textit{time pressure}, testing participants under \textit{time-constrained} and \textit{unconstrained} conditions. To interpret the resulting behaviors, we developed a family of computational models that factorize physical prediction and planning independently. On the prediction dimension, we compare an \ac{ipe} module running probabilistic forward simulations against a visual-heuristic predictor employing a learned classifier (\cref{fig:intro_c}) \citep{lindsay2021convolutional,calabro2025cnn,szegedy2017inceptionv4}. On the planning dimension, we contrast myopic strategies with deliberative lookahead. By fitting resource parameters (\eg, search depth, noise levels) and comparing predicted action sequences across conditions, we find that human physical reasoning reflects resource-rational adaptation at both dimensions: as task demands escalate, people deploy simpler physical prediction mechanisms and adopt shallower planning strategies, balancing cognitive cost against expected reward.

\section{Experimental Design}

\subsection{Participants}

A total of 82 participants (54\% male, 46\% female; mean age $= 21.5 \pm 2.6$) were recruited from an online campus forum. Participants received a base payment plus a performance-proportional bonus. One participant was excluded for failing to pass the practice phase.

\subsection{Stimuli}

The experiment employs a web-based, interactive 2D construction task in which participants arrange a sequence of 6 blocks on a continuous grid of size $8 \times 8$ (\cref{fig:stimuli}b). Blocks are randomly drawn from three distinct shapes defined by dimensions $w \times h$ (width $\times$ height), where $w \in \{0.6, 1.2, 1.8\}$ and $h = 0.6$.

Physical dynamics are simulated via PyBullet \citep{coumans2015bullet} at a temporal resolution of $1/240$ seconds over a 5-second window per sequence. To approximate the visual input of real-world physical reasoning, scenes are rendered into photorealistic 2.5D representations using PyRender (\cref{fig:stimuli}a).

Each trial specifies a predetermined block sequence of length $T = 6$, denoted $(b_1, \dots, b_T)$, fully visible to participants. The environment is defined in the $x$--$z$ plane. After $t$ placements, the tower geometry is represented as
\begin{equation}
    G_t = \big((b_1, \mathbf{p}_1), \dots, (b_t, \mathbf{p}_t)\big), \qquad \mathbf{p}_i = (x_i, z_i),
\end{equation}
where $b_i$ indexes the block type and $\mathbf{p}_i$ is its translational placement on the grid. The first block is pre-positioned at the ground center. The decision state at step $t$ is defined as $s_t := (G_t,\, b_{t+1:T})$, augmenting the current geometry with the remaining blocks to be placed.

At each step $t$, the participant selects an action $a_t = (x, z)$ from the legal action set $\mathcal{A}(s_t)$, where $x$ specifies a continuous horizontal position and $z$ a discrete height layer. The action set enforces two physical constraints: (i) \emph{non-penetration}---no overlap with existing blocks or the ground; and (ii) \emph{support}---the new block must rest on a supporting face of an existing block (\cref{fig:stimuli}c). Tower stability is given by a binary indicator $\mathbb{I}_{\mathrm{stable}}(G_t)$, evaluated by the physics engine.

The objective is to maximize horizontal overhang---the maximum lateral distance of any block from the structural centerline (\cref{fig:stimuli}d):
$
    \Omega(G_T) = \max_{i \le T}\left(|x_i| + \tfrac{1}{2}w_i\right),
$
where $w_i$ is the width of the $i$-th placed block. The episode reward is contingent on continuous stability throughout construction: if the tower remains stable at every step, the participant receives the final overhang; otherwise, the reward is zero:
$
    R = \left(\prod_{k=1}^{T} \mathbb{I}_{\mathrm{stable}}(G_k)\right) \cdot \Omega(G_T).
$

\subsection{Task Characteristics}

Unlike binary satisfaction tasks, the objective of \benchmark creates a continuous trade-off: extending blocks shifts the center of mass toward the edge, progressively depleting the ``stability budget.'' A greedy planning strategy that maximizes immediate overhang rapidly exhausts this budget, rendering myopic strategies suboptimal.

Optimal performance therefore requires non-greedy strategies that sacrifice immediate reward for long-term stability. For instance, \textit{counter-weighting} shifts the center of mass inward, while \textit{vertical anchoring} constructs sandwich-like interlocking structures to clamp extended blocks. Because these tactics demand precise causal ordering---blocks must be sequenced to maintain intermediate stability---they serve as a diagnostic signature of deliberative lookahead, making \benchmark well-suited for dissociating planning from physical prediction.

\subsection{Procedures}

\paragraph{Task interface}
Participants used a 2D mouse interface to position a ``ghost block'' preview (\cref{fig:stimuli}) that provided geometric validity feedback (\eg, collision or support validation) but no physical stability cues. We recorded continuous mouse trajectories and preview dwell times to reconstruct participants' internal search processes.

\paragraph{Familiarization}
Prior to the main experiment, participants completed an interactive tutorial covering the web interface, physics dynamics, and scoring rules, followed by 3 practice trials excluded from the stimulus pool. To advance, participants were required to exceed a minimum performance threshold. Unlike the main task, the practice phase permitted trial-and-error repetition, and the optimal tower configuration was displayed after each trial to provide clear ground truth for the expected strategy.

\paragraph{Experiment}
Participants were randomly assigned to one of two between-subjects conditions: \textit{time-constrained} or \textit{unconstrained}. In the \textit{time-constrained} condition, each placement was subject to a strict 5-second limit; failure to confirm a placement within this window terminated the trial and recorded a reward of zero. In the \textit{unconstrained} condition, participants were free to deliberate for an unlimited duration before each placement. Under both conditions, participants completed 20 experimental trials following familiarization, each presenting a distinct task sequence, for a total session duration of approximately 20 minutes. A comprehensive behavioral log---including action sequences, timestamps, and continuous mouse trajectories---was recorded for subsequent analysis.

\section{Models}

To investigate the cognitive mechanisms underpinning sequential physical planning, we develop a computational framework that factorizes the planning process into two independent components: a physical prediction module that evaluates the stability of candidate configurations, and a planning module that expands the search space before committing to an action.

On the prediction dimension, we contrast an \acl{ipe} that estimates stability via Monte Carlo simulation with a visual-heuristic model implemented as a \ac{cnn} \citep{szegedy2017inceptionv4,calabro2025cnn}. On the planning dimension, we distinguish myopic action selection from deliberative lookahead via bounded forward search. We evaluate these paradigms based on both task performance and alignment with human behavioral data.

\subsection{Prediction Models}

\paragraph{\ac{ipe}}
This paradigm instantiates the physical prediction mechanism via Monte Carlo probabilistic simulation, representing physical stability reasoning as mental simulation under noise. For each candidate action, the model runs $K$ forward simulations with stochastic perturbations to estimate stability, where $\mathrm{Stable}^{(i)}$ denotes whether the $i$-th run remains stable:
\begin{equation}
    \hat{p}_{\text{IPE}}(\mathrm{stable}\mid s_t, a) = \frac{1}{K}\sum_{i=1}^{K}\mathbb{I}\!\left[\mathrm{Stable}^{(i)}(G_{t+1})\right].
\end{equation}
We apply Gaussian perturbations $\mathcal{N}(0, \sigma^2)$ to the position, gravity, and friction of each block. Following \citet{wang2024probabilistic,li2025dual}, we set $K = 50$ and $\sigma = 0.03$ to best align with human predictions. This approach is computationally expensive but captures uncertainty arising from perceptual and dynamic noise.

\paragraph{Visual-heuristic model}
This paradigm approximates physical prediction through perceptual heuristics rather than explicit simulation. Because \ac{cnn} extracts statistical patterns without explicit physical laws---sharing visual feature preferences with human judgments \citep{calabro2025cnn}—we train a network as a tractable proxy for cognitive models given the task's complexity, mapping a rendered image of the post-action geometry to a stability probability:
\begin{equation}
    \hat{p}_{\text{NN}}(\mathrm{stable}\mid s_t, a) = f_\theta(I(G_{t+1})).
\end{equation}
Training data consists of 200k diverse partial tower configurations sampled from the task space and excluded from the stimulus pool, with ground-truth stability labels provided by the physics engine. We employ Inception-V4 \citep{szegedy2017inceptionv4} as the model backbone, which has been shown to excel at stability prediction by integrating multi-scale visual cues \citep{calabro2025cnn}. The model is trained for 15 epochs, reaching 97.5\% accuracy on the validation set.

\subsection{Planning Algorithms}

Both planning algorithms select actions by evaluating candidates according to their expected reward and stability, computed using the internal physical prediction module $\mathcal{P}$. The critical distinction lies in the temporal horizon of evaluation: myopic planning considers only immediate consequences, whereas deliberative lookahead evaluates multi-step futures before committing.

\paragraph{Myopic planning}
The myopic planner selects each action based solely on its immediate expected value, without anticipating how current choices constrain future possibilities. At decision state $s_t$, the planner evaluates each candidate action $a \in \mathcal{A}(s_t)$ by:
\begin{equation}
    a^*_t = \arg\max_{a \in \mathcal{A}(s_t)} \; \hat{p}_{\mathcal{P}}(s_t, a) \cdot r(s_t, a),
\end{equation}
where $r(s_t, a)$ is the immediate reward. This strategy is computationally cheap but systematically neglects downstream consequences.

\paragraph{Deliberative lookahead}
The lookahead planner imagines future trajectories before acting. At each decision state $s_t$, the planner expands candidate action sequences up to depth $D$ and evaluates their utilities:
\begin{equation}
    U(\pi \mid s_t) = \left( \prod_{k=t}^{t+D-1} \hat{p}_{\mathcal{P}}(s_k, a_k) \right) \cdot V(s_{t+D}),
\end{equation}
where $\pi = (a_t, \dots, a_{t+D-1})$ and $V(s_{t+D})$ estimate the overhang of the resulting state (\eg, cumulative overhang at depth $D$). Larger $D$ enables deeper foresight at greater computational cost, sacrificing immediate reward for long-term gain---the behavioral signature of counter-weighting strategies.

\section{Results}

The basic statistics between the two groups are shown in \cref{tab:overall_comparison}. Results from 81 participants indicate that time constraints significantly reduced the overhang of successful solutions ($p < .001$) without compromising the overall success rate ($p = .484$). This pattern suggests that under time pressure, humans adopt a risk-averse planning strategy, sacrificing potential reward to ensure stability.

\begin{figure*}[t!]
    \centering
    \small
    \begin{subfigure}[t]{0.625\linewidth} 
        \centering
        \includegraphics[width=\linewidth]{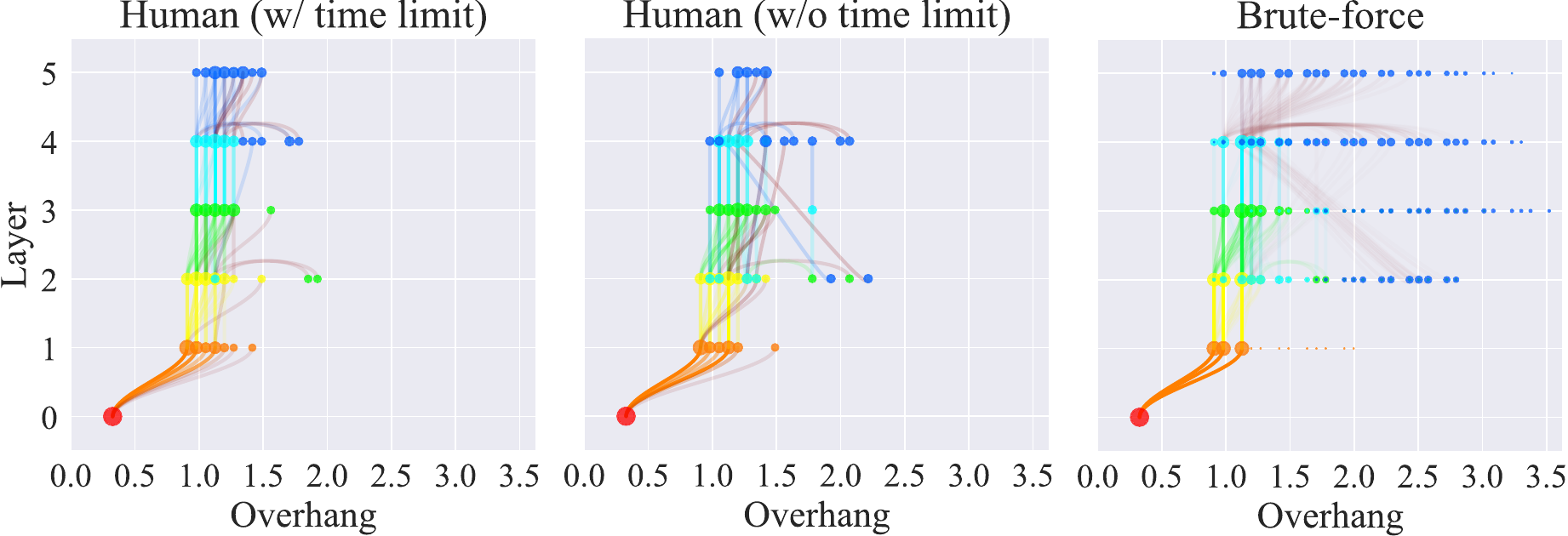}
        \caption{}
        \label{fig:traj_dist}
    \end{subfigure}%
    \hfill%
    \begin{subfigure}[t]{0.371\linewidth}
        \centering
        \includegraphics[width=\linewidth]{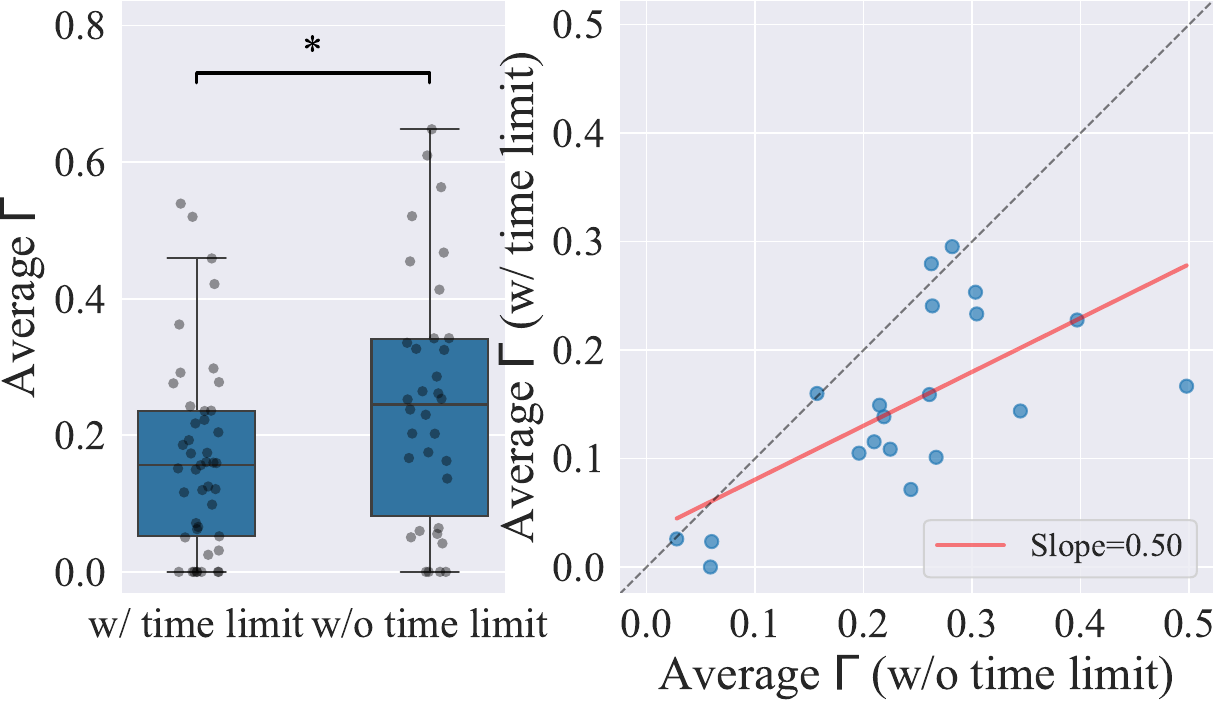}
        \caption{}
        \label{fig:lookahead_gamma}
    \end{subfigure}%
    \caption{\textbf{Planning trajectory distributions and $\Gamma_{G_T}$.} (a) Planning distributions for a block sequence with widths $0.6, 1.8, 1.2, 1.8, 0.6, 1.8$. Flow color encodes stability (native hue = stable; crimson = unstable), opacity indicates normalized transition probability. The near-optimal solution yields overhang 2.4, whereas the best human result was 1.94. Time-constrained participants exhibited a stronger tendency toward vertical stacking; early differences remain modest because both groups first establish a conservative base, with divergence clearer later. (b) Left: distribution of $\Gamma_{G_T}$ across individual participants under each condition. Right: distribution of $\Gamma_{G_T}$ across tasks.}
    \label{fig:lookahead_distribution}
\end{figure*}

\begin{figure}[b!]
    \centering
    \small
    \includegraphics[width=\linewidth]{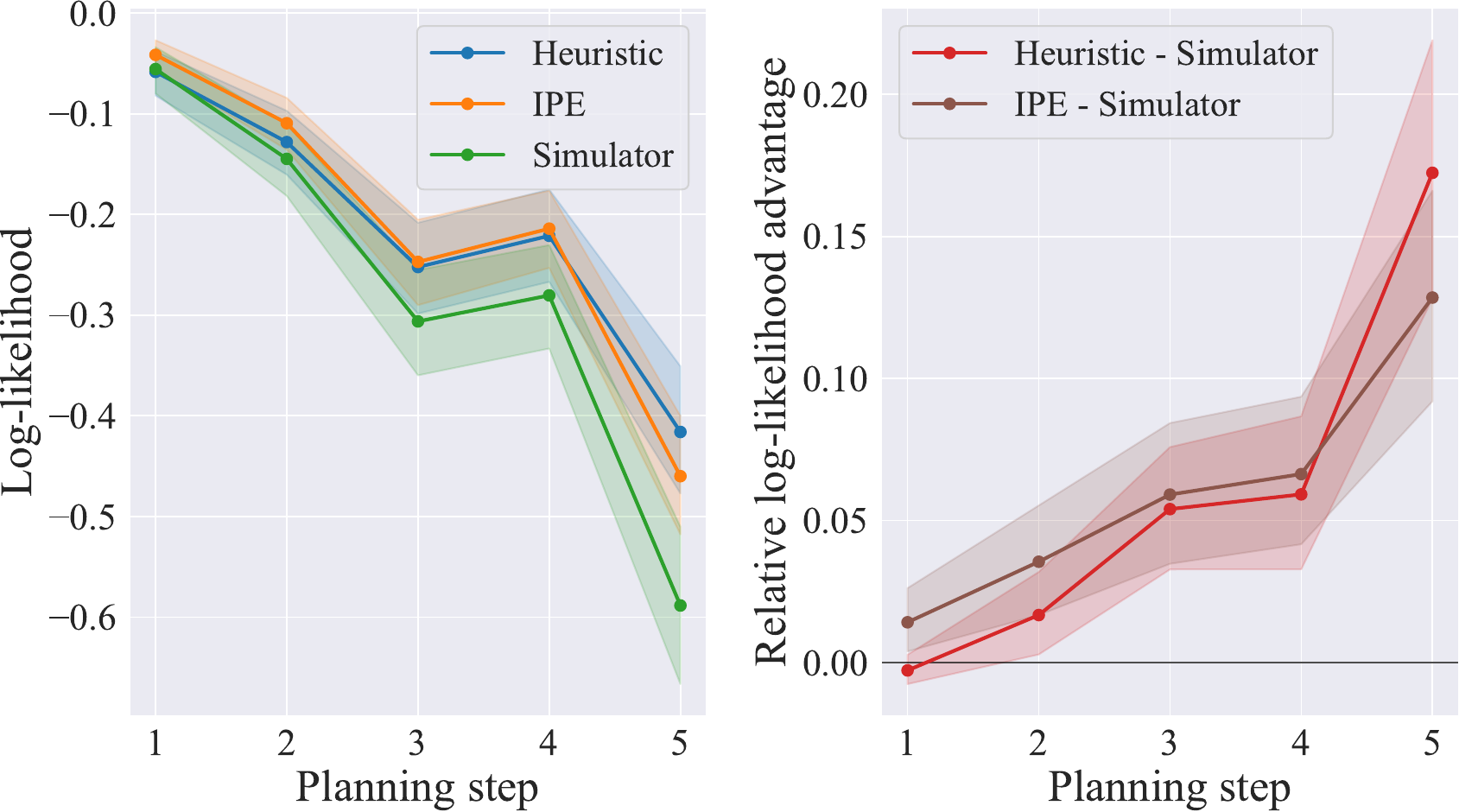}
    \caption{\textbf{Physical prediction mechanism performance across structure complexity.} The left panel shows the log-likelihood of stability predicted by three models on human-rational states as construction progresses; larger values indicate a better fit to human judgments. The right panel shows the relative log-likelihood advantage of \acs{ipe} \vs visual heuristics, computed as the residual with respect to the veridical simulator. As scene complexity grows over successive placements, the visual-heuristic model surpasses \acs{ipe}.}
    \label{fig:model_loglikelihood}
\end{figure}

\begin{table}[b!]
    \small
    \centering
    \setlength{\tabcolsep}{3pt}
    \caption{\textbf{Overall performance.} Comparison between participants in the time-constrained ($n = 43$) and unconstrained ($n = 38$) conditions. Average overhang is computed over successful trials only; decision time indicates the mean decision time per placement. Values indicate mean $\pm$ \acs{sem} unless otherwise stated. Statistical significance assessed using permutation tests.}
    \label{tab:overall_comparison}
    \begin{tabular}{lccc}
        \toprule
        & \multicolumn{2}{c}{Time Limit} & \\
        \cmidrule(lr){2-3}
        Metric & 5s & None & $p$-value \\
        \midrule
        Total reward & $18.25 \pm 0.57$ & $19.57 \pm 0.56$ & .107 \\
        Stable proportion (\%) & $70.2 \pm 2.5$ & $67.8 \pm 2.1$ & .482 \\
        Average overhang & $1.321 \pm 0.023$ & $1.460 \pm 0.029$ & $<.001$ \\
        Decision time (s) & $2.47 \pm 0.07$ & $ 7.27 \pm 0.88$ & $<.001$ \\
        \bottomrule
    \end{tabular}
\end{table}

\subsection{Physical Prediction Mechanism Switching}

To validate our hypothesis concerning physical prediction mechanisms, we compared human behavior against two candidate models---a visual-heuristic model and an \ac{ipe}---with a noise-free veridical simulator serving as baseline.

Assuming stepwise stability judgments and rational action selection, we treat all executed actions as positive samples of perceived physical plausibility, and measure mean log-likelihood across all human actions (\cref{fig:model_loglikelihood}). Both approximate models significantly outperformed the veridical baseline ($p < .001$), confirming that human intuitive physics systematically deviates from Newtonian mechanics---people rely on noise-corrupted representations rather than faithful physical simulation.

More critically, the relative explanatory power of the two physical prediction mechanisms varied across the planning trajectory. In early stages, when the scene was sparsely populated, \ac{ipe} exhibited a modest advantage. As scene complexity grew with additional blocks, however, this advantage reversed: the visual-heuristic model decisively outperformed \ac{ipe} ($p < .001$).

This crossover reflects a fundamental asymmetry in how the two mechanisms scale with structural depth. In \ac{ipe}, perceptual noise---modeled as Gaussian perturbations $\mathcal{N}(0, \sigma^2)$---propagates through forward simulation: positional uncertainty in lower blocks cascades into contact forces above, causing predictive variance to grow with stack height. Stability estimates thus become increasingly diffuse and uninformative for complex towers. If human mental simulation faces an analogous computational bottleneck, continuing to invest limited cognitive resources in a physical prediction mechanism whose reliability rapidly degrades would be resource-irrational.

The visual-heuristic mechanism plausibly sidesteps this complexity penalty by mapping static geometric features---such as vertical misalignment or unsupported overhang---directly onto stability estimates, without unrolling contact dynamics forward in time. Its representational cost therefore remains roughly constant regardless of stack depth. This asymmetry naturally motivates a hybrid physical prediction architecture: human planners invest cognitive effort in deliberate \ac{ipe}-style simulation for simple structures where it is reliable, but shift adaptively to efficient visual heuristics when compounding uncertainty renders deep simulation prohibitive.

\subsection{Behavioral Evidence for Deliberative Lookahead}

We first visualize the planning trajectories of participants in \cref{fig:traj_dist}. Participants under time pressure exhibited a stronger preference for vertical stacking---a conservative planning strategy that prioritizes immediate stability---whereas unconstrained participants explored more laterally extended configurations. This divergence suggests that time pressure suppresses multi-step lookahead. To test this hypothesis more rigorously, we developed a structural metric that operationalizes lookahead in terms of \emph{order dependency}.

Our key insight is that deliberative planning strategies---counter-weighting and vertical anchoring---exploit physical dependencies that emerge only when blocks are placed in a precise causal sequence. A counterweight must be positioned \emph{before} the cantilevered block it stabilizes; reversing this order causes collapse. Thus, if a tower's final geometry can only be achieved through a small fraction of possible construction orders, this path dependency serves as a behavioral signature of deliberative lookahead. Formally, for a completed tower $G_T$, let $\Pi(G_T)$ denote the set of all geometrically valid construction orders---permutations satisfying non-penetration and support constraints at each step. We define \textbf{Order Dependency} as:
\begin{equation}
    \Gamma_{G_T} = 1 - \frac{|\{\pi \in \Pi(G_T) \mid \mathrm{Stable}(\pi)\}|}{|\Pi(G_T)|}.
\end{equation}
High $\Gamma$ indicates that most feasible construction paths fail due to intermediate instability---the structure is path-dependent and requires deliberative sequencing.

As shown in \cref{fig:lookahead_gamma}, time pressure significantly reduced Order Dependency (Unconstrained: $\Gamma = 0.24 \pm 0.03$; Time-constrained: $\Gamma = 0.15 \pm 0.02$; $p = .022$). This difference reveals a qualitative shift in planning strategy: unconstrained participants constructed towers demanding precise sequencing to maintain stability, whereas time-constrained participants retreated to order-invariant heuristics---trading structural complexity for cognitive efficiency.

\begin{table}[t!]
    \centering
    \small
    \setlength{\tabcolsep}{3pt}
    \caption{\textbf{Terminal reward across 20 tasks.} Averaged terminal reward for human participants and computational models. Stochastic models were run 40 times to reduce variance. $D$ indicates lookahead search depth.}
    \label{tab:results}
    \begin{tabular}{lc}
        \toprule
        \textbf{Model} & \textbf{Terminal Reward} \\
        \midrule
        Human (w/ time limit) & $0.913 \pm 0.03$ \\
        Human (w/o time limit) & $0.979 \pm 0.03$ \\
        Myopic & $0.52 \pm 0.12$ \\
        Lookahead ($D=2$) & $0.912 \pm 0.10$ \\
        Lookahead ($D=3$) & $1.180 \pm 0.24$ \\
        \bottomrule
    \end{tabular}
\end{table}

\subsection{Computational Evidence for Deliberative Lookahead}

Building on prior results, we evaluate the joint architecture of sequential physical planning by coupling the stage-dependent hybrid \ac{ipe}-heuristic physical prediction mechanism with varying depths of forward search. Because computing exact action-level likelihoods in our continuous, high-dimensional state space is computationally intractable, we focus on the aggregate structural achievements generated by these coupled models across 20 tasks. The results, alongside human performance, are summarized in \cref{tab:results}.

The results reveal a clear dissociation between myopic and deliberative planning strategies. Guided by the same physical prediction mechanism, the myopic planner achieves a terminal reward drastically below human baselines. This performance gap arises because myopic planning greedily pursues immediate overhang, rapidly exhausting the stability budget---confirming that human physical reasoning cannot be reduced to reactive, single-step utility maximization.

Extending the planning horizon reliably recovers performance. The correspondence between human performance under varying time constraints and models of different search depths suggests that time pressure truncates, but does not eliminate, deliberative lookahead. Taken together, these results demonstrate that successful sequential physical planning requires a tightly coupled architecture: a physical prediction mechanism capable of estimating stability under uncertainty, embedded within a planning strategy that extends deep enough to navigate the risk-reward trade-off.

\section{Discussion}

\paragraph{Coupling between prediction and planning}
We primarily treated physical prediction mechanisms and planning strategies as independent and separable  processes. However, resource-rational agents likely vary both dynamically \citep{callaway2018resource,lieder2020resource}, and their coupling may give rise to mixed or emergent mechanisms. Future work could examine how co-adaptation between these two capacities shapes behavior under graded resource constraints.

\paragraph{Disentangling error sources}
Human physical prediction is calibrated to noisy real-world perception and uncertain dynamics \citep{smith2013sources}, whereas simulators implement idealized deterministic physics \citep{davis2014scope}. Consequently, some errors may reflect mismatches between human uncertainty and simulator assumptions, rather than planning limitations alone. An important future direction is to better dissociate errors from imperfect physical prediction and those from limited planning.

\paragraph{Exploring diverse physical phenomena}
Our study focuses on structural stability, a largely static physical domain. Whether the observed trade-offs between physical prediction mechanisms and planning strategies generalize to dynamic phenomena---such as collision or long-range causality \citep{wang2024probabilistic}---where predictions must be propagated over longer temporal horizons and errors may accumulate over time, remains an open question.

\paragraph{Impact of task exposure}
Currently, our framework evaluates each task independently. However, humans actively adapt through trial-and-error. Future work could examine how cumulative task exposure shapes behavior. Beyond improving performance, environment familiarity and prior experience might systematically shift how participants weigh simulation against heuristics or adjust planning horizons across trials.

\section{Conclusion}

We introduced \benchmark, a construction task in which we manipulated time pressure to test whether humans adaptively shift both prediction mechanisms and planning strategies. Our results support a resource-rational architecture: \ac{ipe}-based simulation dominated early stages while visual heuristics prevailed as complexity grew; concurrently, time pressure truncated deliberative lookahead toward shallower horizons. These findings unify two previously isolated debates---simulation \vs heuristics and myopic \vs deliberative planning---as a dynamic repertoire reconfigured by cognitive budget.

By treating humans as active, capacity-limited agents, our framework offers a foundation for understanding physical problem-solving. Future work could extend these findings to dynamic environments and explore whether AI agents, which currently exhibit notable shortcomings in physical reasoning \citep{li2022learning}, can leverage this flexible architecture to achieve human-like physical reasoning.

\paragraph{Acknowledgement}

The authors would like to thank Ruizhe Chen (Peking University), Bingrui Gong (Peking University), and Yuchen Lin (Peking University) for their helpful discussions. This work is supported in part by the Brain Science and Brain-like Intelligence Technology---National Science and Technology Major Project (2025ZD0219400), the National Natural Science Foundation of China (62376009), the PKU-BingJi Joint Laboratory for Artificial Intelligence, the Wuhan Major Scientific and Technological Special Program (2025060902020304), the Hubei Embodied Intelligence Foundation Model Research and Development Program, and the National Comprehensive Experimental Base for Governance of Intelligent Society, Wuhan East Lake High-Tech Development Zone.

\balance
\printbibliography

\end{document}